# Impact of a Batter in ODI Cricket Implementing Regression Models from Match Commentary

Ahmad Al Asad
Department of Computer Science and and Engineering
Brac University
Dhaka, Bangladesh
ahmad.al.asad@g.bracu.ac.bd

Kazi Nishat Anwar
Department of Computer Science and Engineering
Brac University
Dhaka, Bangladesh
kazi.nishat.anwar@g.bracu.ac.bd

Ilhum Zia Chowdhury
Department of Computer Science and Engineering
Brac University
Dhaka, Bangladesh
ilhum.zia.chowdhury@g.bracu.ac.bd

Akif Azam
Department of Computer Science and Engineering
Brac University
Dhaka, Bangladesh
akif.azam@g.bracu.ac.bd

Tarif Ashraf
Department of Computer Science and Engineering
Brac University
Dhaka, Bangladesh
tarif.ashraf@g.bracu.ac.bd

Tanvir Rahman
Computer and Information Sciences
College of Engineering
University of Delaware
Newark, DE, USA
rtanvir@udel.edu

*Abstract*—Cricket, a "Gentleman's Game," is a prominent sport rising worldwide. Due to the rising competitiveness of the sport, players and team management have become more professional with their approach. Prior studies predicted individual performance or chose the best team but did not highlight the batter's potential. Our research, on the other hand, aims to evaluate a player's impact while considering his control in various circumstances. This paper seeks to understand the conundrum behind this impactful performance by determining how much control a player has over the circumstances and generating the 'Effective Runs,' a new measure we propose. We first gathered the fundamental cricket data from open source datasets; however, variables like the pitch, weather, and control were not readily available for all matches. As a result, we compiled our corpus data by analyzing the commentary of the match summaries. This gave us an insight into the particular game's weather and pitch conditions. Furthermore, ball-by-ball inspection from the commentary led us to determine the control of the shots played by the batter. We collected data for the entire One Day International career, up to February 2022, of 3 prominent cricket players: Rohit G Sharma, David A Warner, and Kane S Williamson. Lastly, to prepare the dataset, we encoded, scaled, and split the dataset to train and test Machine Learning Algorithms. We used Multiple Linear Regression (MLR), Polynomial Regression, Support Vector Regression (SVR), Decision Tree Regression, and Random Forest Regression on each player's data individually to train them and predict the Impact the player will have on the game. Multiple Linear Regression and Random Forest give the best predictions accuracy of 90.16% and 87.12%, respectively.

## I. INTRODUCTION

Cricket has become a popular sport among different age groups globally. One of the major formats of this game is "One Day International (ODI)," a 50-over format of the game. The pitch and weather conditions added to the dynamics of the opposition are always a challenge for the players in this particular format. One of the crucial aspects of the game is the pitch, which alone can dictate the match's outcome. Furthermore, the weather conditions are equally important as the conditions may not always be suitable for a batter. A batter deserves to be acclimated whenever they perform sublimely in adverse conditions, thus having an "Impact" on the game's outcome. The impact, in this case, reflects not only the runs scored but also the amount of "Control" they have had. This control is a better representation of the impact of the performance. Most of the past works are used to predict player performance [15] or identify the best team [16]. In contrast, our research seeks to assess a player's influence while considering his control in all sorts of situations. This paper aims to decipher the dilemma behind this impactive performance through our research on how much control a player has over the situation and using different Machine Learning algorithms and compare their results to generate "Effective Runs." To summarize, our proposal is unique in that it considers the control of batters in measuring of the player's impact on the overall match.

The rest of the paper is organized as follows: Section II. describes the Literature Review. Section III describes the Methodology used for generating the preliminary dataset. Section IV explains our Experimental Setup and Section V shows the Implementation of algorithms and the data visualization. Finally, Section VI is the Conclusion followed by Future Scope, Acknowledgement and References.

## II. LITERATURE REVIEW

In this current world of entertainment, irrespective of the sport in context, classification, and ranking of players has become an important factor for many and has thus brought to life a vast area of interest for many researchers. This chapter of the paper aims at proving the importance of research in the field of cricket through reviewing previous relevant works in the field. A group of researchers developed a new player rating system. This proposed contribution method could then rate players over time based on their batting, bowling, and fielding contributions and determine the best player in a match, a series, or a calendar year [1]. Later in 2017, another couple of authors, incorporated game situations and the team strengths to measure the player contributions and named it the "Work Index," representing the amount of work that is yet to be done by a team to achieve its target [2]. They authenticated their approach with an 86.80% accuracy for predicting the player of the match in ODI. In [3], the authors proposed a model to predict the performance of batters using a multilayer perceptron (MLP) neural network which gave them an accuracy of 66.67%. This was then used to select the players to include in a team. Another group of researchers analyzed how factors like home game advantage, day/night effect, winning the toss, and batting first affect the outcome of the match using Bayesian classifiers by developing a software tool CricAI that can increase the chances of victory through simple tweaks in certain factors [4]. However, using statistical analysis, it was concluded that there is no competitive advantage in winning the coin toss and also the log-odds of the probability of winning increases by around 50 percent when playing on one's home field [5]. Another way of predicting performance was implemented by classifying cricketers into three categories - performer, moderate and failure. They progressively trained and tested their neural network models, and obtained results favorable to the World Cup 2007 [6]. Moreover in [7], the authors focused on additional factors like pitch type, weather, ground, opposition to build a model for predicting the player's performance and finding the best all-rounder. Their model gave the best outcome with Random Forest after using a number of machine learning algorithms. The authors in [8] determined that high individual wickets, number of bowled deliveries, number of the thirties, total wickets, wickets in the power play, runs in death overs, dots in middle overs, and number of fours and singles in middle overs impose a high impact on the result of a game using SVM with an 81% accuracy. Besides, another group of authors created a range of variables [9]. Their model used the prediction variables which were numerically weighted according to statistical significance and used to predict the match outcome. In [10], Logistic Regression identified factors that play a key role in predicting ODI match outcomes. Later, the authors used SVM and Naïve Bayes Classifier for model training and predictive analysis and determines the following variables: home-field advantage, winning the toss, game plan (batting or fielding first), match type (day or day & night), competing team, venue familiarity, and season the match is played in. While in another research, the importance of the opposition faced was further described by talking into account the runs scored by the batter the bowler is bowling against by making a new measure for individual performance called "Quality" [11].

We are proposing a new measure called "Impact" which is the Effective Run scored by a batter taking into account the runs scored and the control the batter had on the balls faced. We used Machine Learning algorithms to predict this effective run using corpus features like pitch type, opposition, weather and other extra features.

III. METHODOLOGY

The work plan first involves collecting the readily available data. Then we analyzed the commentary for individual games and generated more data manually. Some of them were indexed to give different weights, while the others used to derive new data. Next step was to generate a measure that can represent the impact of a batter. Regression Models were then used to predict this impact.

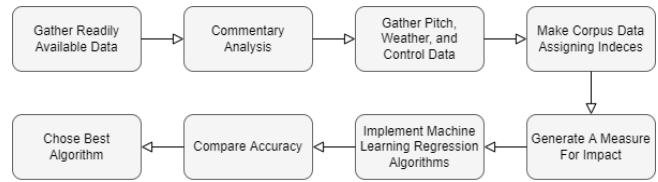

Fig. 1. Work plan of the research.

*A. Primary Dataset*

We did our preliminary data analysis by collecting data for the batting records of more than 500 matches throughout the individual careers of Rohit Sharma, David Warner, and Kane Williamson. Espncricinfo has its database for cricketers called Statsguru. Here [12], we collected the readily available data for: "Bat1," "Runs," "BF," "SR," "4s," "6s," "Opposition," "Ground," and "Start Date." Another data source, Cricmetric [13] contained an additional "Dot Ball %" data for all the matches. These data were complemented by collecting some extra data which were not readily available: "Win/Loss," "Team Runs," and "In At Position Number."

The aforementioned variables were then used to derive some more important features. "Bat1" consisted of data on whether the batter played in a game or not, and also contained data on whether they were dismissed in that game, hence the variable "Out/Not Out." "Ground" contained the venue data and hinted at the data for "Home/Away." The "Dot Ball %" was used to get the "Dot Ball" data, which in turn gave the "Scoring Shot" data which indicated the number of deliveries the batter used to score a run and that was used to make a new variable "Scoring Rate" which is Runs per Scoring Shot. Finally, the data for "Others" were generated using the "Runs," "4s," and "6s" to give the "Running between the wickets fraction."
**The columns in the combined preliminary dataset are** - "Bat1," "Out/NotOut," "Runs," "BF," "SR," "4s," "6s," "Opposition," "Ground," "Home/Away," "Start Date," "Dot Ball %," "Dot Ball," "Scoring Shot," "Scoring Rate," "Others," "Running between the wickets fraction," "Win/Loss," "Team Run," "In at Position number," etc.

## B. Corpus Data Input

The second stage of data collection required us to make our corpus data. The "Opposition" column contained the teams played against. We proposed a quantifiable way to weigh each team. A batter's performance also majorly depends on the type of "Pitch" played on and the "Weather" of the day. Furthermore, our proposed system focuses hugely on the "Control" of the batter throughout the game. This control was manually sought out by analyzing the commentary of individual deliveries.

Firstly, a batter's capability is judged by their performance against the top-rated ODI teams. We accumulated the ICC ODI team ranks for each opposition team for each match. Next, we assigned an index to each rank, designed based on the threat level teams set forth for batters. However, the team rankings fluctuate every so often and it is challenging to differentiate the weight between consecutive ranks like $2^{nd}$ and $3^{rd}$ or $7^{th}$ and $8^{th}$ etc. To reduce this disparity, the team ranks were divided into groups of 3. The top 3 ranked teams pose an equal threat, for which we assigned them a weight of "5". Similarly, "4" is given to teams ranked from 4 to 6, "3" given to teams ranked 7-9, "2" for teams ranked 10-12, and "1" for teams ranked 13 and below.

Secondly, cricket pitches are generally categorized into a green top, a dry/dusty pitch, and a dead/flat track [17]. Similar to the opposition above, a batter's performance varies considerably with different pitches. A flat track, for example, is the most suitable for the batter to score. In contrast, a green top is a bowler-friendly pitch and is the most challenging for a batter to perform. A batter showing high performance in a green top indicates that the batter has more control over the deliveries he faces. A dry pitch tests the batter as well as gives them an equal opportunity to perform. Hence, we encoded the pitches as follows - "Green" : 2, "Dry" : 1.5, "Flat" : 1. The primary source of the pitch details was the commentary and post-match presentation. In some situations with no details, assumptions were made regarding the nature of the pitch based on the general location and venue.

Next, weather condition has been derived into four categories: "Clear," "Sunny," "Windy," and "Overcast." The adaptability of a batter in any of these categories will define their performance. The weather information was collected from the pre-match analysis commentary where the initial weather report was considered. Although there had been some ambiguity, the data was kept mostly consistent in regards to some keywords such as: "hot and humid" for sunny, "breezy" for windy, and "cloudy and dark skies" for overcast.

Lastly, control corresponds to a batter's gracefulness and finesse in hitting playable deliveries from the middle part of the bat, these being "middled" shots, alongside refusing to strike a good delivery, these being "left alone" shots [14].

$$\text{Control} = \frac{\text{Middled Shots} + \text{Left Alone Shots}}{\text{Total Balls Faced}} \quad (1)$$

Espncricifno generates the control of batters and only makes it publicly available given a noteworthy performance in a match. We had 53 available control statistics for Rohit, and the remaining had to be done manually. Initially, we analyzed 29 of the 53 available matches and compared our calculated control with the actual control. The comparison is expressed in a graph shown in Figure 2.

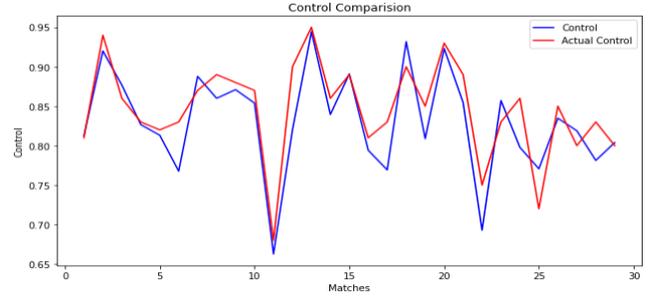

Fig. 2. Comparison of Actual Control and Calculated Control.

Our calculated control has a 3.24% uncertainty compared to the actual control, thus proving our commentary analysis to be highly accurate with low uncertainty. Following this process, we analyzed match commentary for every delivery for the three players' matches. We tallied all the middled and left alone shots, excluding the edged shots and not appropriately connected with the bat. For left alone, some common keywords used were: "ducked," "stepped away," " no shot offered," "shoulders arm," and many more. For middled, we searched for some keywords such as: "defended," "swayed," "drives firmly and straight," "controlled," and various other adjectives that defined excellence.

Fig. 3. Commentary data sample from Espncricinfo.

## IV. EXPERIMENTAL SETUP

We have introduced a new parameter of a player's performance that incorporates both runs and control of a batter, eventually leading to the "Effective Runs" scored. This is what we are calling "Impact."

### A. Effective Runs

For devising the Effective Runs, our initial approach was to generate a heatmap similar to Figure 4, that would give the correlation between the features. Upon analyzing the heatmap, the Control was observed to vary in contrast to the other variables, similar to Runs.

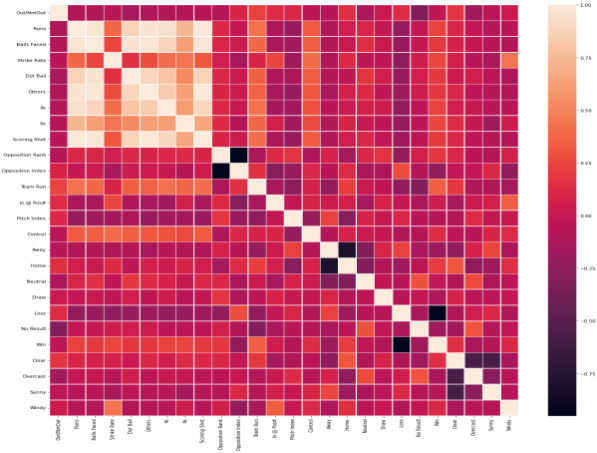

Fig. 4. Correlation heatmap of variables over David A Warner's ODI career.

In a regular game of cricket, the Runs Scored by the batter were the main focus. We propose that the Control of a batter is also a variable that can be defined as a score for the batter, alongside Runs. The more the control, the better the batter has a hold in the game's direction. Hence, we defined the Effective Runs as an exponential relationship between Runs Scored and Control, as in Equation 2. We are referring to this Effective Runs as Impact.

$$\text{Effective Runs} = \text{Runs Scored} * e^{\text{Control}} \quad (2)$$

**The final dataset includes the previous dataset along with the following new columns** - "Opposition Rank," "Opposition Index," "Middled," "Left Alone," "Control," "Actual Control," "Pitch," "Pitch Index," "Weather," "Impact," etc.

**The completed datasets for all three players is available here** - https://tinyurl.com/odi-batter-dataset.

### B. Data Preprocessing

For preprocessing, the entire dataset was imported using the pandas dataframe. The first step was to discard data with missing values. Next, for dimension reduction, some variables were discarded. Variables like Runs, Control, Strike Rate, etc. that had a direct connection with the Effective Runs and other variables like Opposition, which were used to derive a new weight, were also discarded.

**The specific columns included as features are** - "Out/NotOut," "Opposition Index," "Home/Away," "Dot Ball," "Others," "Win/Loss," "Team Run," "In at Positon number," "Pitch Index," "Weather," etc. The label was "Impact."

The categorical non-numerical features such as, "Out/NotOut", "Home/Away", "Win/Loss", and "Weather" were encoded. Label Encoding was used for "Out/NotOut" with 0 for "Out" and 1 for "NotOut," whereas one-hot encoding was used for the remaining three features. The numerical features had to be scaled for better performance of the Machine Learning algorithms. This was achieved by using the *StandardScaler()* function from the sklearn library.

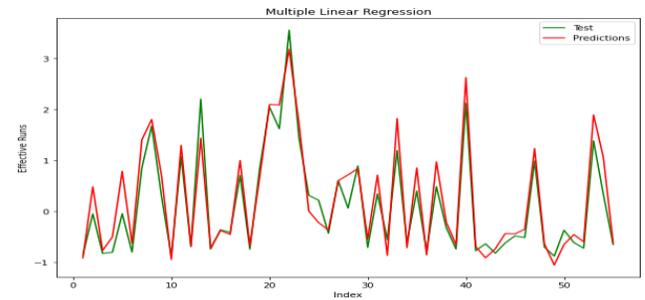

Fig. 5. Part of the Preprocessed Dataset.

## V. IMPLEMENTATION

The machine learning algorithms used were all based on regression. The features selected as independent variables were used to determine the "Impact" and the results were compared to determine the best algorithm.

The first step was to split the data into test and train datasets using the function *train_test_split()* in a 1:3 ratio. Support Vector Regression, Polynomial Regression, Multiple Linear Regression, Decision Tree Regression, and finally Random Forest Regression were all trained individually using the train dataset for each of the three players. Ultimately, these trained models were used to predict the dependent variable for the test dataset.

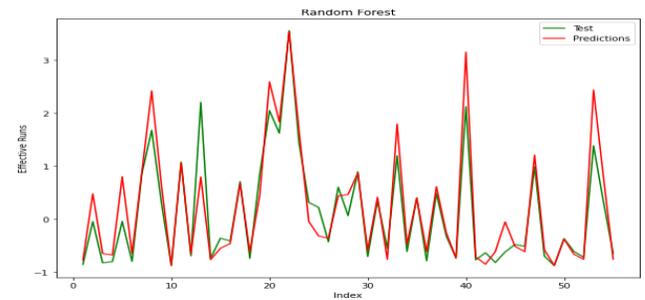

Fig. 6. Comparison of Actual and Predicted Effective Runs using Multiple Linear Regression for Rohit G Sharma.

Fig. 7. Comparison of Actual and Predicted Effective Runs using Random Forest Regression for Rohit G Sharma.

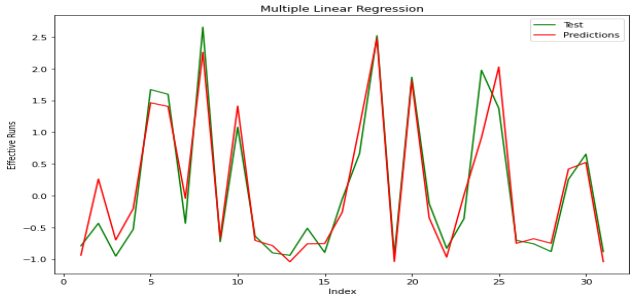

Fig. 8. Comparison of Actual and Predicted Effective Runs using Multiple Linear Regression for David A Warner.

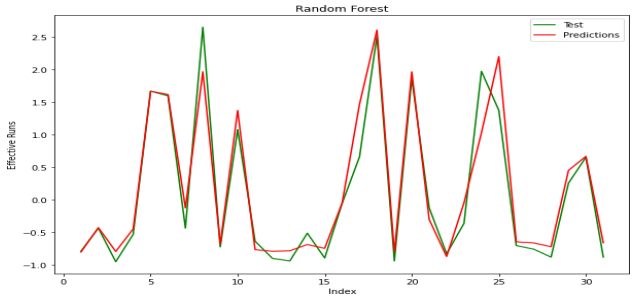

Fig. 9. Comparison of Actual and Predicted Effective Runs using Random Forest Regression for David A Warner.

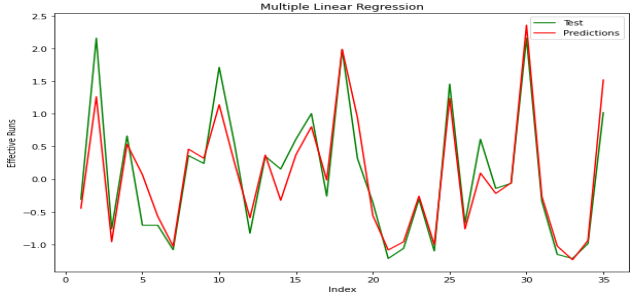

Fig. 10. Comparison of Actual and Predicted Effective Runs using Multiple Linear Regression for Kane S Williamson.

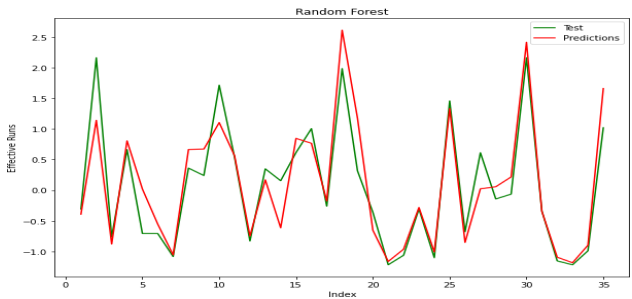

Fig. 11. Comparison of Actual and Predicted Effective Runs using Random Forest Regression for Kane S Williamson.

Another method from sklearn library, *r2_score()* function was used to compare and calculate the accuracy of the predicted values with the label for the test dataset. The library matplotlib was then used for data visualization as figures 6–11 show how the actual test dataset label varies from the predicted values.

*A. Results Analysis*

Our proposed measure considers both the Runs Scored and the Control over the Balls Faced by the batter. Upon comparing the predicted values with the actual test dataset, the accuracy of each player's individually trained models is showcased in Table I.

The Polynomial Regression model gave the weakest prediction. Polynomial Features from sklearn library was used with *degree = 4* which an accuracy of 76.07% for Sharma, 70.01% for Warner, and 56.34% for Williamson. On average, an accuracy of 67.47% was observed with this model. For Support Vector Regression, the kernel was set as 'rbf,' giving accuracies of 79.88%, 80.21%, and 73.13% distributed amongst the players. This result averaged 77.74%.

Decision Tree Regressor was used with *random_state = None*. This gave an accuracy of 79.37% which averaged from the individual 74.72%, 90.00%, and 73.39%. But, upon using Random Forest Regressor with *n_estimators = 10*, building ten trees significantly improved the results. Average accuracy of 87.12% was observed from this trained model. Figures 7, 9 and 11 show how the predicted and actual data varies.

TABLE I
ACCURACY FROM PREDICTIONS.

| Regression Model | *Sharma* (%) | *Warner* (%) | *Williamson* (%) |
|:---:|:---:|:---:|:---:|
| Multiple Linear | 89.14 | 91.53 | 89.80 |
| Random Forest | 85.06 | 91.70 | 84.60 |
| Support Vector | 79.88 | 80.21 | 73.13 |
| Polynomial | 76.07 | 70.01 | 56.34 |
| Decision Tree | 74.72 | 90.00 | 73.39 |

However, the best outcome was reached using Multiple Linear Regression. As pictured in figures 6, 8 and 10, our set of features - using the proposed method for weighing the opposition in terms of their rank, and weighing the pitch in terms of effortlessness of play and the categorization of weather - will give their best outcomes in predicting the Effective Runs, with an average accuracy of 90.16%. This Effective Runs can hence be used to measure a batter's impact in a game. This impactive performance can give the team selectors and even the players themselves an idea of how their gameplay can deviate the direction a match takes and make the sport even more interesting.

## VI. CONCLUSION

Cricket has successfully settled its area of research. As the game continues to develop, the passion for the game has illuminated the youth and has given an exponential rise to potentially excellent cricketers. Discovering the effectiveness of the control of a batter significantly contributes to the outcome of a cricket match. It helps the team management to recognize better players with match-winning capabilities. The research began by collecting and processing data on the set of features that highly affect a batter's performance. The control of a batter was the main focus of making a new measure to determine how their performance can change the flow of a game. Features like pitch, weather, opposition and other extra factors were used along with Machine Learning models to predict the Impact of a batter in an ODI match. Multiple Linear Regression gave the best results with an accuracy of 90.16%. This new measure, "Effective Runs," can be used to determine the impactive performance of a batter.

## FUTURE SCOPE

Our Future agenda is to tune our dataset further and propose new models that may even give better accuracy. In the future, our motive is to implement Artificial Neural Network (ANN) for our models. We are planning to make new measures for batters in Test and T20 cricket as well; additionally, our motive is to create new measures for bowlers and allrounders.

## ACKNOWLEDGEMENT

We would like to convey our gratitude to late Mr. Hossain Arif, Assistant Professor, CSE, Brac University, for his constant guidance, which has considerably aided us in moving forward with our research. We express our heartfelt condolences on our professor's untimely passing.